\documentclass[11pt]{article}
\RequirePackage{silence}
\usepackage{amssymb}
\usepackage{longtable}
\usepackage{booktabs}
\usepackage{booktabs}
\usepackage{makecell}
\usepackage{array}
\usepackage{ragged2e}
\usepackage{tabularray}
\UseTblrLibrary{booktabs}

\usepackage{listings}
\usepackage{xcolor}
\usepackage{subcaption}

\usepackage[preprint]{acl}
\usepackage{times}
\usepackage{latexsym}
\usepackage{blindtext}
\usepackage{amsmath}
\usepackage[T1]{fontenc}
\usepackage{tcolorbox}
\usepackage{xcolor}
\usepackage{enumitem}
\usepackage{amssymb}
\usepackage{booktabs}
\usepackage{multirow}
\usepackage{soul}

\usepackage[utf8]{inputenc}

\usepackage{microtype}

\usepackage{inconsolata}

\usepackage{graphicx}

\title{Are Arithmetic Heuristic Neurons Form-Invariant? A Mechanistic Analysis of Symbols, Text, and Code in LLMs}

\author{
    Sharath Naganna\textsuperscript{1,*,\dag} \quad
    Tanvir Ahmed Sijan\textsuperscript{2,*} \quad
    \textbf{Uddipta Kalita}\textsuperscript{3} \\
    \\
    \textsuperscript{1} Indian Institute of Technology Bombay, Mumbai, India\\
    \textsuperscript{2} Jahangirnagar University, Dhaka, Bangladesh \\
    \textsuperscript{3} National Institute of Technology, Silchar, India\\
    \texttt{sharathhn@cse.iitb.ac.in}\quad
\texttt{\{sijantanv, uddiptain20\}@gmail.com}  \\
\textsuperscript{*} Equal contribution \quad \textsuperscript{\dag} Corresponding author
}

\usepackage{lipsum}
\begin{document}
\maketitle
\begin{abstract}
Large language models often succeed on one formulation of a problem while failing on an equivalent formulation. Whether these failures arise from distinct internal circuits or different activation states of a shared circuit remains unknown. Recent mechanistic interpretability studies suggest that arithmetic in LLMs emerges from a "bag of heuristics," encoded by a sparse set of MLP neurons that represent distinct arithmetic strategies. We investigate whether arithmetic heuristic neurons are form-invariant across symbolic arithmetic, natural language word problems, and Python code in three Llama-3 models. In each format, we identify arithmetic heuristic neurons using a two-stage pipeline combining attribution patching and activation patching. A compact set of neurons is shared across all three formats, and targeted interventions show this shared circuit is both necessary and sufficient for late-layer arithmetic computation. Transferring the shared neurons' activations from a successful execution in one format to a failed execution in another recovers most incorrect predictions, exceeding 97\% for addition and subtraction, indicating that cross-format failures arise from activation states rather than distinct circuits. Moreover, shared neurons consistently belong to the same heuristic families across formats, demonstrating that arithmetic computation in LLMs is largely form-invariant at the neuron level.
\end{abstract}

\section{Introduction}
Large language models (LLMs) have demonstrated remarkable mathematical capabilities, achieving impressive performance on challenging arithmetic and mathematical reasoning benchmarks \citep{cobbeTrainingVerifiersSolve2021,hendrycksMeasuringMassiveMultitask2020,weiChainofThoughtPromptingElicits2023}. These advances have established LLMs as powerful mathematical problem solvers, capable of tackling problems ranging from elementary arithmetic to graduate-level mathematics. Consequently, understanding how these models internally represent and perform mathematical computation has become an important research direction, motivating a growing body of work in mechanistic interpretability aimed at uncovering the neural mechanisms underlying arithmetic reasoning \citep{stolfoMechanisticInterpretationArithmetic2023b,wuPyveneLibraryUnderstanding2024,marjiehWhatNumberThat2025}.

\begin{figure}[t]
    \centering
\includegraphics[width=\columnwidth]{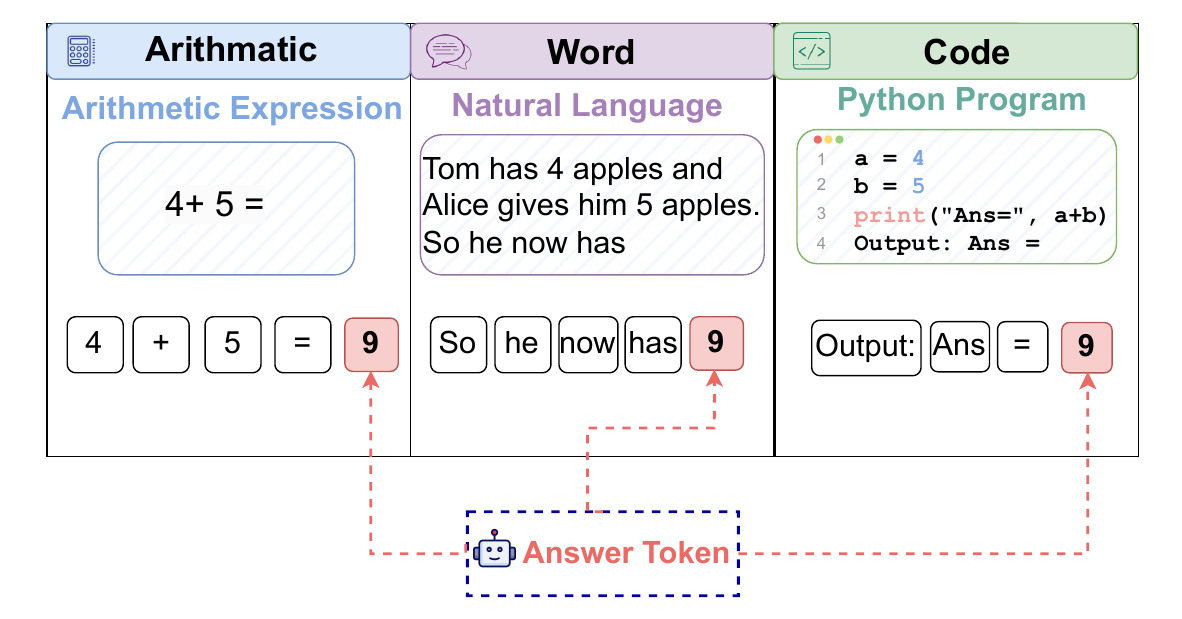}
    \caption{
    Three prompt formats used in our experiments. Each expresses the same arithmetic computation as symbolic arithmetic, Python code, or a natural language word problem. The answer is always generated as a single token at the final position.}
    \label{fig:prompt_formats}
\end{figure}

Recent mechanistic interpretability studies have begun identifying the internal neural circuits responsible for arithmetic computation in LLMs. In particular, \citet{nikankinArithmeticAlgorithmsLanguage2025} demonstrate that LLMs solve arithmetic problems using a ``bag of heuristics,'' where distinct neurons encode different arithmetic strategies and can be causally manipulated to influence model predictions. While these findings provide compelling evidence that arithmetic behavior emerges from identifiable neural mechanisms, they are derived almost exclusively from canonical symbolic expressions \citep{baeumelModularArithmeticLanguage2025, duFineGrainedManipulationArithmetic2025,nikankinArithmeticAlgorithmsLanguage2025}. Consequently, it remains unclear whether the same arithmetic neurons generalize across alternative representations of the same computation or are specific to a particular input format.

This question is particularly important because previous studies have shown that LLMs are highly sensitive to the formulation of mathematical problems, where even small changes to a problem's surface form, such as paraphrasing, adding new statements, or altering the presentation while preserving the underlying computation, can lead to substantial performance degradation \citep{liGSMPlusComprehensiveBenchmark2024}. LLMs also exhibit noticeable variance in their performance when responding to different instantiations of same problem \citep{mirzadehGSMSYMBOLICUNDERSTANDINGLIMITA2025}.

From a mechanistic perspective, however, the source of this sensitivity remains poorly understood. While recent studies have shown that LLMs separate problem abstraction \citep{chengCanLLMsReason2025} and operand routing \citep{mamidannaAllOneLLMs2025} from the final computation, it remains unclear whether the same neuron-level arithmetic heuristics are recruited across fundamentally different input formats, namely symbolic equations, word problems, and code.

Motivated by this gap, we investigate the following research questions:
\begin{itemize}
    \item \textbf{RQ1: Are arithmetic heuristic neurons form-invariant?}
    Do mathematically equivalent problems expressed as symbolic equations, word problems, and code recruit the same arithmetic heuristic neurons?

    \item \textbf{RQ2: What causes cross-format failures?}
    Do performance differences across formats arise from distinct arithmetic circuits or different activation states of a shared circuit?
\end{itemize}
Code and data for reproducing our experiments are available.\footnote{\url{https://github.com/SharathHN/format-invariant-arithmetic}}


\begin{figure*}[t]
    \centering
    \includegraphics[width=0.95\textwidth]{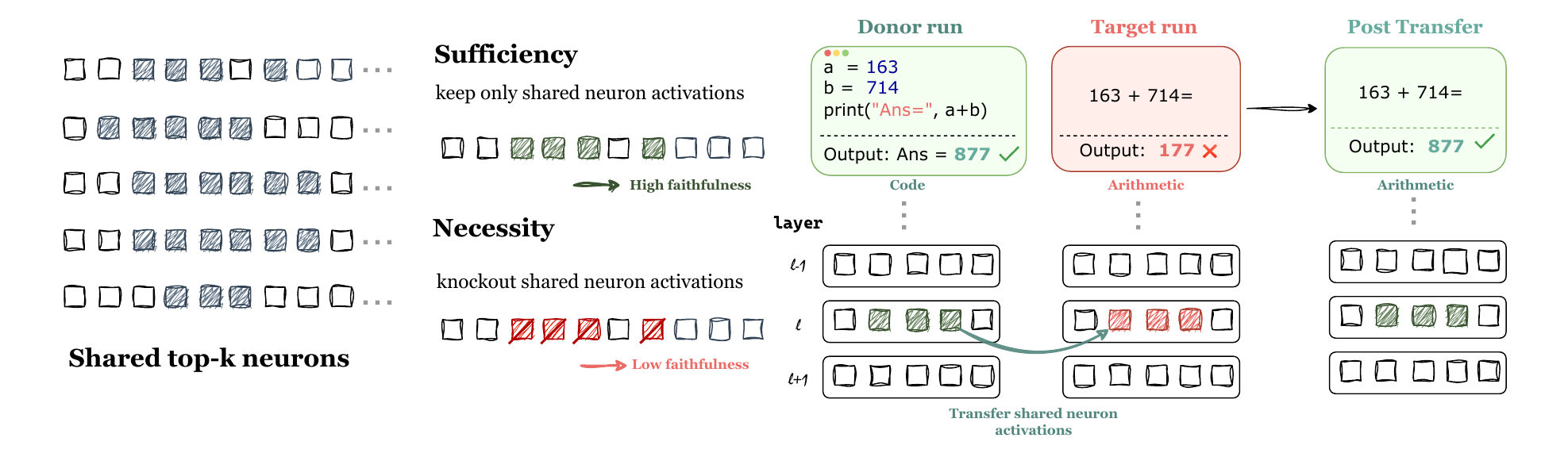}
    \caption{\textbf{Overview of the pipeline.} (Left) We identify the top-k causal neurons independently for symbolic arithmetic, word problems, and code, computing their layer-wise intersection to isolate the shared circuit. (Middle) We evaluate whether these shared neurons are sufficient and necessary for arithmetic computation via Keep-Only (retaining only shared activations) and Knockout (ablating only shared activations) interventions. (Right) We transfer the activations of the shared neurons from a successful donor execution to a failed target execution, testing whether cross-format failures stem from differences in activation states rather than distinct arithmetic circuits.}
    \label{fig:overview}
\end{figure*}

\section{Related Work} 
\subsection{Mechanistic Interpretability}
Mechanistic interpretability (MI) aims to reverse-engineer the internal mechanisms of language models by analyzing their learned weights and computational components. In transformer-based LLMs, computations are often understood in terms of circuits, small sets of interacting components, such as MLP neurons and attention heads, that collectively implement a specific function \citep{baeumelModularArithmeticLanguage2025,gouldSUCCESSORHEADSRECURRING2024}. To identify these circuits, mechanistic interpretability relies heavily on causal mediation techniques \citep{pearlDirectIndirectEffects2013}, which localize model behavior to specific internal components. Representative methods include activation patching \citep{mengLocatingEditingFactual2023, zhangCarefulExaminationLarge}, path patching \citep{goldowskydillLocalizingModelBehavior2023,wangInterpretabilityWildCircuit2022}, attribution patching \citep{nanda2023attributionpatching,syed-etal-2024-attribution,hanna2024have}, and logit attribution \citep{nostalgebraist2020logitlens,belroseElicitingLatentPredictions2025}. 

\subsection{Arithmatic Reasoning Interpretability}
Recent mechanistic interpretability studies have begun to uncover how LLMs perform arithmetic reasoning. Circuit analyses have identified specialized model components that play causal roles in arithmetic computation \citep{stolfoMechanisticInterpretationArithmetic2023b,zhangInterpretingImprovingLarge2024}, while subsequent work suggests that arithmetic emerges from collections of heuristic neurons \citep{nikankinArithmeticAlgorithmsLanguage2025} rather than a single coherent algorithm \citep{nandaProgressMeasuresGrokking2023, NEURIPS2023_56cbfbf4, dingSurvivalFittestRepresentation2024}. Alternative explanations propose that LLMs rely on symbolic pattern matching \citep{dengLanguageModelsAre2024}, dual-pathway computation separating coarse and fine-grained numerical processing \citep{lindsey2025biology}, or geometric representations of numbers that support arithmetic operations \citep{kantamneniLanguageModelsUse2025}. More recently, \citet{chengCanLLMsReason2025} showed that LLMs solve arithmetic word problems through an abstract-then-compute mechanism, in which models first construct an abstract representation of the underlying arithmetic operation before executing the computation. They further showed that these abstract representations are invariant to different natural language formulations of the same problem. Relatedly, \citet{mamidannaAllOneLLMs2025} identified a sparse attention subgraph that routes operands to the last token position for direct arithmetic. However, they found this subgraph fails on word problems and code, suggesting that while the core routing mechanism generalizes across simple text prompts, complex formats require additional computational components for deeper semantic and syntactic parsing.

Building upon this line of work, we investigate whether arithmetic heuristic neurons themselves are form-invariant. Specifically, we systematically identify and causally validate arithmetic heuristic neurons to determine whether the same neurons are recruited to perform equivalent computations across different formats, namely symbolic arithmetic, word problems, and code.

\section{Background}
\subsection{MLPs as Key-Value Memories}

Transformer layers consist of two primary components: a multi-head self-attention module, which routes information across tokens, and a Multi-Layer Perceptron (MLP), which performs nonlinear transformations independently at each token position. Following the key-value memory interpretation of \citet{gevaTransformerFeedForwardLayers2021,gevaTransformerFeedForwardLayers2022}, the MLP can be expressed as

\begin{equation}
\mathbf{y}=\sigma(\mathbf{x}\mathbf{K})\mathbf{V},
\end{equation}

where $\mathbf{x}\in\mathbb{R}^{d}$ and $\mathbf{y}\in\mathbb{R}^{d}$ denote the input and output representations, $\mathbf{K}\in\mathbb{R}^{d\times d_{\mathrm{mlp}}}$ and $\mathbf{V}\in\mathbb{R}^{d_{\mathrm{mlp}}\times d}$ are the up-projection and down-projection weight matrices, respectively, and $\sigma(\cdot)$ is a nonlinear activation function. Omitting the bias terms, the MLP can be decomposed into individual neurons as

\begin{equation}
\mathbf{y}
=
\sum_{i=1}^{d_{\mathrm{mlp}}}
\sigma(\mathbf{x}\cdot\mathbf{K}_{:,i})\mathbf{V}_{i,:}
=
\sum_{i=1}^{d_{\mathrm{mlp}}}
H_i\mathbf{V}_{i,:},
\end{equation}

where

\begin{equation}
H_i=\sigma(\mathbf{x}\cdot\mathbf{K}_{:,i})
\end{equation}

denotes the activation of the $i$-th neuron. Under this formulation, each neuron is represented by a key vector $\mathbf{K}_{:,i}$ and a corresponding value vector $\mathbf{V}_{i,:}$. The key vector determines the neuron's activation by measuring how well the input matches its learned pattern, while the value vector specifies the information written back to the residual stream. Consequently, the MLP output is a weighted sum of the value vectors of all activated neurons.

\subsection{Arithmetic Neurons and Heuristics}

Building upon the neuron interpretation of MLPs, recent mechanistic interpretability studies have identified a sparse subset of neurons that play a causal role in arithmetic computation \citep{nikankinArithmeticAlgorithmsLanguage2025,duFineGrainedManipulationArithmetic2025}. These \emph{arithmetic neurons} are primarily located in the middle and later layers of Transformer models and are selectively activated during arithmetic reasoning. Let $M$ denote a language model and $S_{\mathrm{arith}}$ denote the set of arithmetic neurons identified within the model. Prior work shows that ablating these neurons, producing a modified model $M \setminus S_{\mathrm{arith}}$, causes arithmetic performance to deteriorate dramatically while leaving many non-arithmetic capabilities comparatively unaffected. 

Although arithmetic neurons collectively enable numerical reasoning, they do not all perform the same computation. Instead, \citet{nikankinArithmeticAlgorithmsLanguage2025} show that these neurons implement a diverse collection of arithmetic heuristics, where individual neurons respond to specific arithmetic patterns and may participate in multiple heuristic strategies. These heuristics operate over either the operands or the expected result and include \textbf{range} (values within a numerical interval), \textbf{modulo} (specific modular congruencies), \textbf{pattern} (digit patterns), \textbf{identical operands} ($op_1 = op_2$), and \textbf{multi-result} (promoting a small set of division outputs). They describe arithmetic computation as emerging from a ``bag of heuristics,'' in which many specialized neurons cooperate to produce the final prediction.
\section{Experimental Setup}
\label{sec:setup}
We conduct all experiments on three base Llama models: \textbf{Llama-3-8B}, \textbf{Llama-3.2-3B} and \textbf{Llama-3.2-1B}. We select these models because their tokenizers represent every integer in the range $[0,999]$ as a single token, enabling precise neuron-level analysis without complications arising from multi-token numerical representations. We present results for Llama-3-8B in the main paper. Corresponding results for Llama-3.2 3B and 1B exhibit the same qualitative trends and are provided in Appendix \ref{app:additional_results}.

\paragraph{Prompt Formats.}
Following \citet{nikankinArithmeticAlgorithmsLanguage2025}, we construct semantically equivalent arithmetic problems in three input formats: \textbf{symbolic arithmetic}, \textbf{Python code}, and \textbf{natural language word problems} as shown in figure \ref{fig:prompt_formats}. Each prompt expresses the same underlying computation $a \mathbin{\mathrm{op}} b$ over the four arithmetic operators $\{+, -, \times, \div\}$, where division follows Python's integer floor semantics.

    


Although the three formats differ substantially in sequence length, every prompt is constructed so that the numerical answer is generated as a \emph{single token} at the final position. Consequently, all mechanistic analyses are performed on the residual stream and MLP activations at the final (answer) token, providing a consistent basis for cross-format comparison.

\paragraph{Dataset Construction.}
All prompts are zero-shot and \emph{model-conditioned}. Following prior mechanistic interpretability studies \citep{wang2023interpretability,prakash2024finetuning,nikankinArithmeticAlgorithmsLanguage2025}, we retain only operand pairs that the model answers correctly, allowing subsequent analyses to isolate the circuits responsible for successful arithmetic computation. For each operator and input format, operand pairs are sampled to maximize the diversity of unique numerical answers, yielding $200$ correctly solved prompts. These are split into $100$ \emph{training} prompts, used to estimate activation statistics and neuron importance, and $100$ \emph{evaluation} prompts, reserved exclusively for causal intervention and faithfulness experiments.

Division is a special case because valid single-token integer quotients are far less diverse than for the other operators, causing the number of unique answers to saturate (approximately 96, 163, and 164 for arithmetic, code, and word problems, respectively). We therefore interpret them with caution.





\subsection{Layer Identification}
\label{sec:setup:layers}

To identify where arithmetic computation emerges, we train a format-specific, operator-invariant linear probe on the residual stream (\texttt{resid\_post}) at every layer and token position to predict the numerical answer.Each probe is trained on roughly 9,000-–10,000 correctly answered prompts for each model-–format pair using an 80/20 train–test split (see Appendix~\ref{app:dataset_statistics} for dataset statistics).

Across all formats, the answer first becomes reliably linearly decodable at the \emph{final} input token position for Llama 3 8B, with a consistent computation-onset layer. We additionally train pooled probes shared across all four arithmetic operators, which similarly recover a common answer direction within each format. Based on these results, we restrict subsequent neuron analyses to the final (answer) token and to layers $\ell \geq 16$, capturing both the emergence of arithmetic computation and its execution.

\begin{figure}[t]
    \centering
    \includegraphics[width=\linewidth]{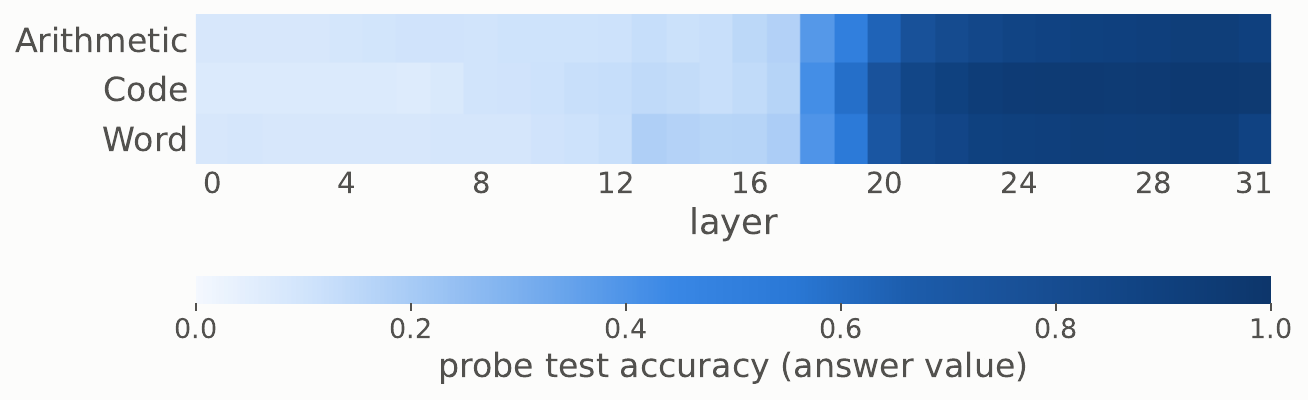}
    \caption{Linear probe accuracy across transformer layers. Arithmetic answers become reliably linearly decodable at a consistent layer across symbolic arithmetic, code, and word problems, motivating our focus on the final token and late-layer MLP neurons.}
    \label{fig:probe}
\end{figure}

\subsection{Identifying Arithmetic Neurons}
\label{sec:setup:ranking}

Having identified the computation-onset layers, we next rank neurons according to their contribution to arithmetic computation. Neuron importance can be estimated using either \emph{gradient-based} methods, such as attribution patching \citep{syed-etal-2024-attribution,hanna2024have,cheng2026drivesrepresentationsteeringmechanistic}, or \emph{gradient-free} methods, such as activation patching \citep{wang2023interpretability,conmy2023towards,zhang2024towards}. Gradient-based approaches are computationally efficient but provide only first-order approximations, whereas gradient-free methods estimate causal effects exactly but are substantially more expensive.

To balance efficiency and accuracy, we adopt a two-stage ranking procedure. We first use attribution patching to identify the top $2000$ candidate neurons per layer, then rerank these candidates using exact activation patching. This substantially reduces computational cost while preserving the ranking quality of gradient-free interventions.

\paragraph{Stage 1: Attribution patching.}

Following \citet{stolfoMechanisticInterpretationArithmetic2023b,nikankinArithmeticAlgorithmsLanguage2025}, neuron importance is quantified using the indirect effect

\begin{equation}
    \text{IE} =
    \frac{1}{2}
    \left(
    \frac{P^{*}(r')-P(r')}{P(r')}
    +
    \frac{P(r)-P^{*}(r)}{P^{*}(r)}
    \right),
    \label{eq:ie}
\end{equation}

where $P$ and $P^{*}$ denote the output probabilities before and after intervention. Attribution patching approximates this quantity using a first-order Taylor expansion,

\begin{equation}
\text{IE}
\approx
(a^{\mathrm{corrupt}}-a^{\mathrm{clean}})
\cdot
\frac{\partial m}{\partial a},
\end{equation}

where $m$ denotes the indirect-effect objective evaluated on the clean run. For each (model, format, operator), importance scores are averaged across the training prompts, and neurons are ranked by $\mu+\sigma$ (mean plus one standard deviation) to prioritize neurons with both high and consistent causal effects.

\paragraph{Stage 2: Activation patching.}

We then rerank the top $2000$ candidates per layer using exact activation patching. For each candidate neuron, its last-token activation is replaced with the corresponding activation from a matched corrupt prompt, and Eq.~\ref{eq:ie} is evaluated exactly using the resulting output probabilities. Average indirect effects over the training prompts are used to obtain the final ranking, from which the top $200$ neurons per layer are retained for all subsequent analyses.

\subsection{Circuit Faithfulness Evaluation}
\label{sec:setup:faithfulness}

We evaluate whether the identified circuit is both \emph{necessary} and \emph{sufficient} for arithmetic computation using a faithfulness metric based on mean ablation. Given a candidate circuit, every MLP neuron \emph{outside} the circuit is replaced, at every token position, with its corresponding activation from a precomputed mean cache. Attention heads and all MLP neurons in layers preceding the computation-onset region ($\ell < 16$) remain unmodified. We then measure how much of the correct-answer signal is preserved under this intervention.

Following \citet{syed-etal-2024-attribution}, we quantify performance using the normalized logit
\[
\mathrm{nl}=\frac{\mathrm{logit}[r]}{\max(\mathrm{logit})},
\]
where $r$ denotes the correct answer token. We evaluate three conditions: (i) the unmodified model (\emph{good}), (ii) complete mean ablation of all attention heads and MLP neurons (\emph{bad}), and (iii) mean ablation of only the non-circuit MLP neurons (\emph{circuit}). Circuit faithfulness is defined as the fraction of the performance gap recovered by the retained circuit:

\begin{equation}
\mathrm{Faithfulness}
=
\frac{\mathrm{nl}_{\mathrm{circuit}}-\mathrm{nl}_{\mathrm{bad}}}
{\mathrm{nl}_{\mathrm{good}}-\mathrm{nl}_{\mathrm{bad}}}.
\end{equation}

\paragraph{Mean Cache.}

Mean ablation requires a neutral replacement activation for every model component. We construct a \emph{mean cache} from a large, evaluation-disjoint set of operand pairs by averaging MLP neuron and attention-head activations at each token position. Because prompts differ in length across formats, activations are aligned relative to the final (answer) token before averaging. During ablation, the cached activations are applied using the same alignment, ensuring that each token position is replaced with statistics computed from comparable positions across prompts.



\subsection{Cross-Format Activation Transfer}
\label{sec:setup:xformat}



To investigate why mathematically equivalent problems are sometimes solved correctly in one format but not another, we perform cross-format activation patching on the shared arithmetic heuristic neurons. For each operator and donor$\rightarrow$target format pair, we identify operand pairs that are solved correctly in the donor format but incorrectly in the target format, evaluating up to 600 matched pairs per transfer direction.

We replace the shared neuron activations at the final (answer) token in the failed target execution with those from the corresponding successful donor execution, leaving all other activations unchanged. We compare three settings: \emph{matched}, where donor and target share the same operands; \emph{mismatched}, where activations come from a different operand pair; and \emph{random}, where an equally sized set of randomly selected neurons is patched. We report the fraction of corrected target predictions and the normalized logit of the correct answer.

\subsection{Neuron Heuristic Families}
\label{sec:setup:families}

To characterize \emph{what} the identified neurons compute, we apply the heuristic-classification framework of \citet{nikankinArithmeticAlgorithmsLanguage2025} to the shared neurons. For each neuron, the method quantifies how consistently its highest-activating inputs follow an interpretable arithmetic pattern, such as operand range, result modulus, or result value. Following the original authors' approach, we evaluate both \emph{indirect} and \emph{direct} heuristics, identifying operand-related features from raw activations and result-related features from activations projected to the vocabulary space. Unless otherwise stated, we use the recommended matching threshold of $0.6$. 

For each arithmetic operator, we classify the neurons in the three-way shared circuit independently for the arithmetic, code, and word-problem formats. This allows us to examine whether neurons shared across formats also exhibit consistent functional roles by implementing the same heuristic family. We quantify this using the mean pairwise Jaccard similarity between family sets and the exact family-match rate across formats. The distribution is shown in Appendix \ref{app:family_distribution}.

\section{Results}
\label{sec:results}

We present evidence that arithmetic computation is implemented by a shared set of form-invariant heuristic neurons from four complementary perspectives. We first quantify the \textbf{sparsity} of the shared circuit across symbolic arithmetic, code, and word problems. We then establish its \textbf{causal} role through \textbf{necessity} and \textbf{sufficiency} analyses. Next, we investigate whether failures in one format can be repaired using activations from a successful execution in another format. Finally, we examine whether the shared neurons implement consistent arithmetic heuristics across formats.

\subsection{Sparsity: A compact arithmetic circuit is shared across formats}
\label{sec:results:sparsity}

We first ask how much overlap exists between the arithmetic circuits identified independently from symbolic arithmetic, code, and word problems. If arithmetic relied on a large pool of generic neurons, the overlap between the three formats would be expected to increase monotonically as progressively larger circuits are considered.

\begin{figure}[t]
    \centering
    \includegraphics[width=0.7\linewidth]{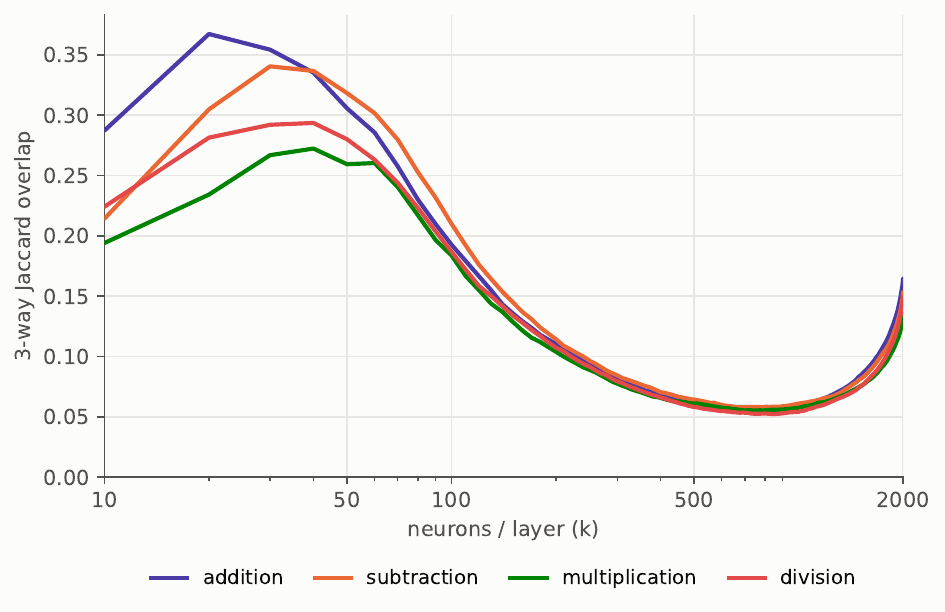}
    \caption{Three-way Jaccard overlap of the top-$k$ neurons across symbolic arithmetic, code, and word problems as a function of circuit size $k$.}
    \label{fig:jaccard_3way}
\end{figure}

Instead, Figure~\ref{fig:jaccard_3way} exhibits a consistent non-monotonic pattern across all arithmetic operators. The three-way Jaccard overlap peaks when approximately $50$--$100$ neurons per layer are retained, before gradually decreasing as additional neurons are included. This suggests that the highest-ranked neurons form a compact core circuit that is consistently recruited across formats, whereas neurons ranked lower become increasingly format-specific. Although the absolute overlap varies by operator, the same qualitative behavior is observed for addition, subtraction, multiplication, and division.

\subsection{Causality: Shared neurons are necessary and sufficient within the late-layer MLPs}
\label{sec:results:causality}

Neuron overlap alone does not establish that the shared neurons are responsible for arithmetic computation. To evaluate their causal role, we compute the three-way intersection of the top-$k$ neurons identified independently from the three formats and perform two complementary interventions.

In the \emph{keep-only} experiment, all neurons outside the shared circuit are mean-ablated, testing whether the shared circuit alone is sufficient for arithmetic computation. Conversely, the \emph{knockout} experiment ablates only the shared circuit while leaving the remainder of the network intact, testing whether the shared neurons are necessary.

\paragraph{Sufficiency.}

\begin{figure*}[t]
    \centering
    \includegraphics[width=\linewidth]{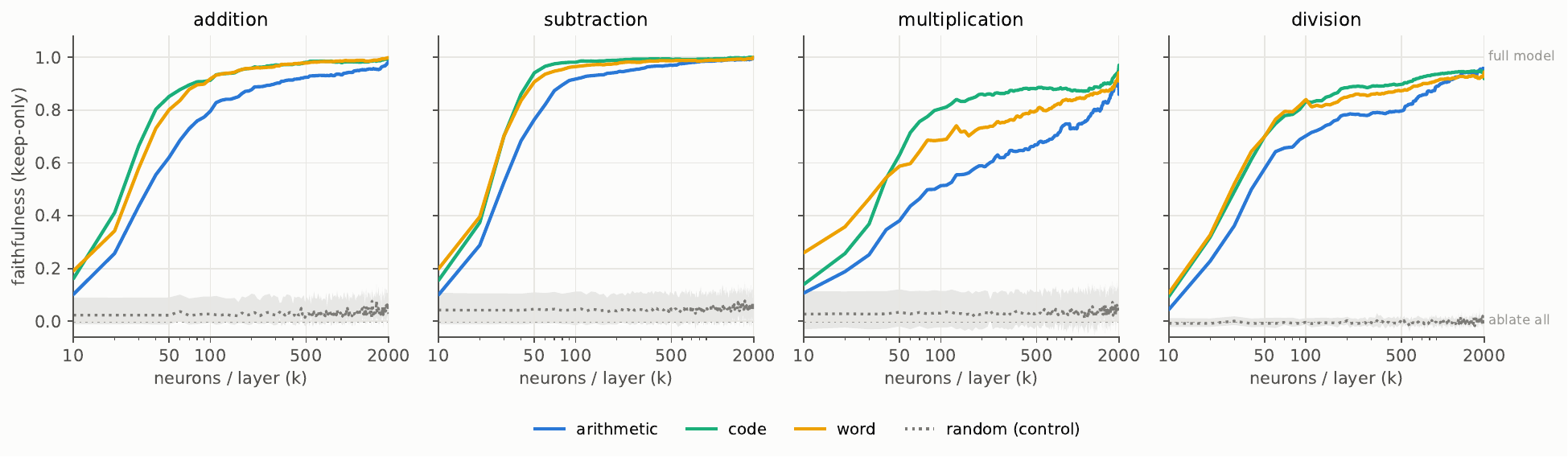}
    \caption{Sufficiency analysis using keep-only interventions. Faithfulness is measured after ablating all neurons outside the shared circuit. Random-neuron controls are matched for circuit size.}
    \label{fig:faithfulness_intersection}
\end{figure*}

Figure~\ref{fig:faithfulness_intersection} shows that the shared circuit rapidly recovers model performance as the circuit size increases. For addition and subtraction, faithfulness approaches the full-model ceiling with approximately $100$--$500$ neurons per layer, while multiplication and division exhibit a more gradual increase. Across all operators and formats, the shared circuit consistently outperforms matched random-neuron controls, demonstrating that a relatively small shared circuit is sufficient to recover nearly all arithmetic performance.

\paragraph{Necessity.}

\begin{figure*}[t]
    \centering
    \includegraphics[width=\linewidth]{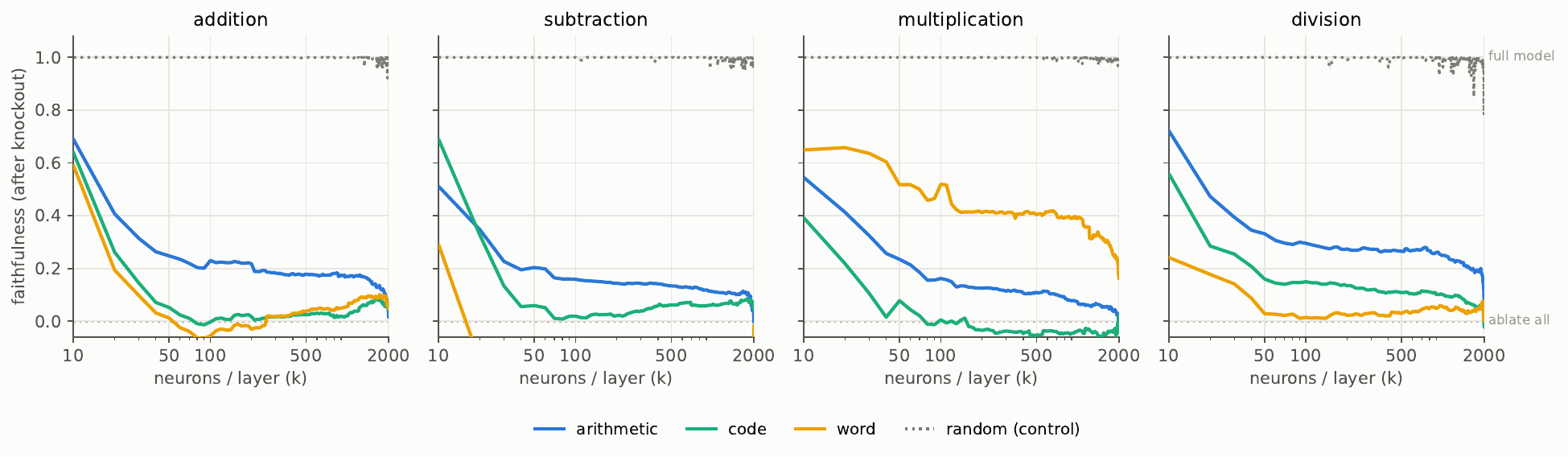}
    \caption{Necessity analysis using knockout interventions. Faithfulness is measured after ablating only the shared circuit. Random-neuron controls are matched for circuit size.}
    \label{fig:faithfulness_knockout}
\end{figure*}

The complementary knockout experiment (Figure~\ref{fig:faithfulness_knockout}) shows that removing only the shared circuit causes a rapid degradation in faithfulness, whereas ablating an equally sized random set of neurons has little effect. 

Together, these interventions demonstrate that the neurons shared across symbolic arithmetic, code, and word problems are both causally necessary and sufficient for arithmetic computation within the late-layer MLPs, while keeping the attention mechanism and earlier layers intact.

\subsection{Cross-format activation transfer recovers most incorrect predictions}
\label{sec:results:repair}

The previous experiments establish that arithmetic computation relies on a shared set of causal neurons across formats. We next ask whether cross-format failures arise from differences in activation states rather than distinct arithmetic circuits.

\begin{figure}[t]
\centering
\includegraphics[width=\linewidth]{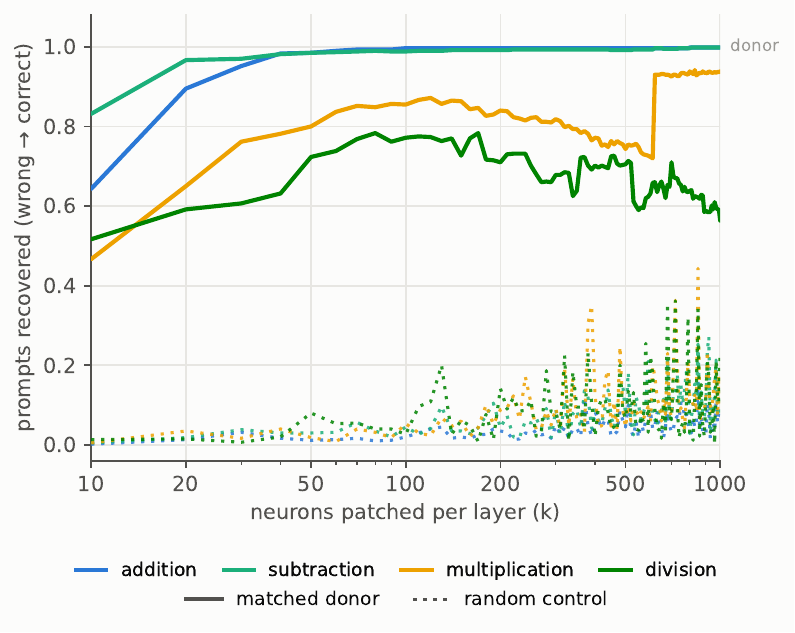}
\caption{Cross-format activation transfer from code to arithmetic using three-way shared neurons. Matched donor activations recover most failed predictions, while random-neuron controls remain near zero.}
\label{fig:code2arith}
\end{figure}

\begin{figure}[t]
\centering
\includegraphics[width=\linewidth]{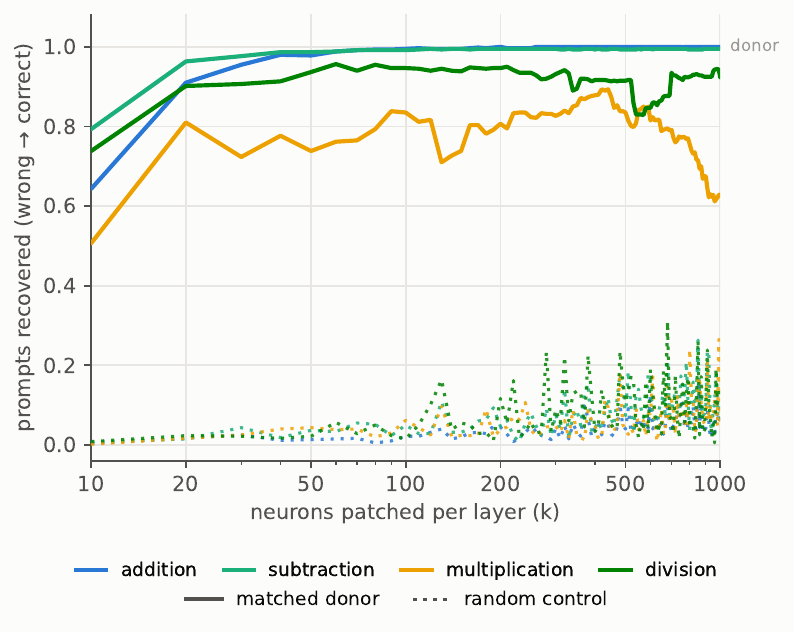}
\caption{Cross-format activation transfer from word
problems to arithmetic using three-way shared neu-
rons. Recovery remains high across operators, whereas
random-neuron controls show negligible improvement}
\label{fig:word2arith}
\end{figure}

Figure~\ref{fig:code2arith} and ~\ref{fig:word2arith} shows the recovery rates for code$\rightarrow$arithmetic and word$\rightarrow$arithmetic transfer across all four operators. Transferring activations from matched donor executions substantially recovers previously incorrect predictions: addition and subtraction exceed 97\% recovery at approximately $k=50$ neurons per layer, while multiplication recovers 80--85\% of failures across both transfer directions. Division also exhibits strong recovery for word$\rightarrow$arithmetic ($\sim$90\%) but more variable behaviour for code$\rightarrow$arithmetic, with recovery declining at larger values of $k$. This suggests that larger intersections increasingly include neurons with code-specific activation patterns that interfere with arithmetic readout. In contrast, patching an equally sized set of randomly selected neurons yields near-zero recovery across all operators and transfer directions, confirming that recovery is specific to the shared heuristic neurons.

These results indicate that cross-format failures primarily arise from differences in the activation states of a shared arithmetic circuit rather than from the recruitment of distinct circuits. Restoring the shared-neuron activations is sufficient to recover most failed predictions.


\subsection{Mechanistic consistency: Shared neurons implement the same heuristics}
\label{sec:results:heuristics}

Across all formats, shared neurons are consistently assigned to the same heuristic families (e.g., operand-range detectors, result-value neurons, and modular arithmetic heuristics). We quantify this agreement using the \emph{exact match rate} (the fraction of neurons assigned the same heuristic family across all formats), \emph{pairwise Jaccard similarity} (the average overlap in heuristic assignments between every pair of formats), and \emph{three-way Jaccard similarity} (intersection-over-union of heuristic assignments across all three formats). As shown in Table~\ref{tab:heuristic_summary}, the pairwise and three-way Jaccard similarities average 0.75 and 0.64, respectively, indicating substantial agreement. Permutation tests confirm that these similarities significantly exceed chance ($p < 10^{-4}$). Although exact agreement is more stringent (30.6\% on average), nearly all shared neurons (94--100\% for addition, subtraction, and multiplication) receive a heuristic label in every format. Full statistics and heuristic-family distributions are provided in Appendices~\ref{app:family_distribution}--\ref{app:heuristic_significance}.

\begin{table}[t]
\centering
\small
\begin{tabular}{lccc}
\toprule
Operator & Exact (\%) & Pair Jaccard & 3-way Jaccard \\
\midrule
Addition & 53.9 & 0.83 & 0.75 \\
Subtraction & 23.5 & 0.73 & 0.62 \\
Multiplication & 20.1 & 0.76 & 0.65 \\
Division & 24.8 & 0.67 & 0.54 \\
\midrule
\textbf{Average} & \textbf{30.6} & \textbf{0.75} & \textbf{0.64} \\
\bottomrule
\end{tabular}
\caption{Functional consistency of shared Llama-3-8B arithmetic neurons across formats. Exact denotes the fraction of shared neurons assigned the same heuristic family in all three formats, while Pair and 3-way denote pairwise and three-way Jaccard similarity of heuristic-family assignments.}
\label{tab:heuristic_summary}
\end{table}


\section{Conclusion}

We investigated whether arithmetic heuristic neurons in LLMs are invariant to the surface form of mathematically equivalent problems. Across symbolic arithmetic, word problems, and code, we found that arithmetic computation relies on a compact set of shared neurons that are both necessary and sufficient for correct computation. Cross-format activation patching showed that restoring their activation states recovers most failed predictions, indicating that cross-format failures arise mostly from differences in activation state rather than distinct circuits. Heuristic-family analysis further showed that these neurons preserve consistent functional roles across representations. Together, these findings demonstrate that arithmetic computation in LLMs is largely form-invariant at the neuron level and mechanistic circuits identified in one representation can generalize across input modalities. Future work may extend this analysis to more complex reasoning, such as algebraic manipulation and multi-step problem solving.



\section*{Limitations}

Our study focuses on elementary integer arithmetic in base Llama models and three carefully controlled input formats. While this setting enables precise causal analysis, it remains unclear whether the observed form invariance extends to more complex mathematical reasoning, instruction-tuned models, or broader language understanding tasks.

Our causal interventions and heuristic analyses focus exclusively on MLP neurons at the final input token. This focus is motivated by the key-value memory interpretation of MLPs, which suggests that they are the primary components responsible for storing and retrieving arithmetic-related representations. Although prior work suggests that attention heads play a complementary role by routing operand information to this position, we leave investigating whether these routing mechanisms also exhibit cross-format invariance to future work.

Finally, we identify shared neurons using prompts that are answered correctly by the model, allowing us to isolate the circuits underlying successful computation. While our cross-format activation transfer experiments examine cases where this shared circuit succeeds in one format but fails in another, extending this analysis to broader reasoning failures and more diverse, unconstrained prompt distributions remains an important direction for future work.

\bibliography{custom}

\clearpage
\twocolumn
\appendix

\appendix

\section{Additional Results on Llama Models}
\label{app:additional_results}

The main paper presents detailed analyses for Llama-3-8B. This appendix reports the corresponding experiments for Llama-3.2-3B and Llama-3.2-1B. Across all analyses, both models exhibit the same qualitative trends as the primary model, supporting the robustness of our findings across model scales.

\subsection{Layer Identification}

\begin{figure}[t]
    \centering
    \includegraphics[width=\linewidth]{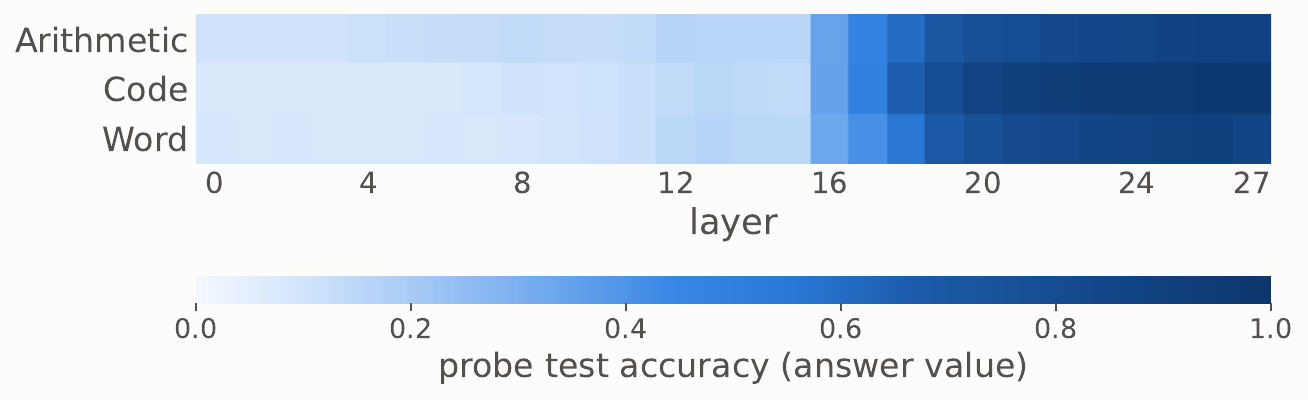}
    \caption{Layer-wise probe accuracy for Llama-3.2-3B.}
    \label{fig:probe_3b}
\end{figure}

\begin{figure}[t]
    \centering
    \includegraphics[width=\linewidth]{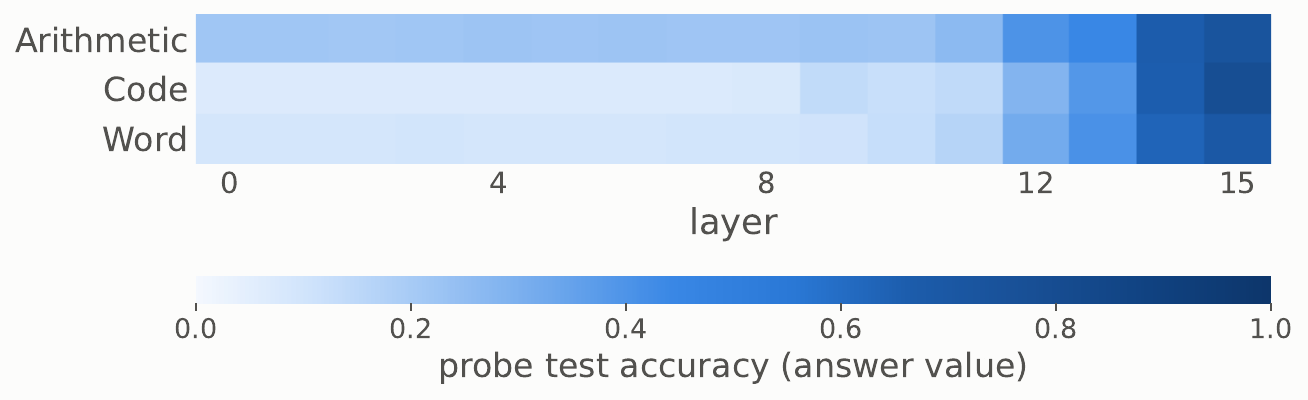}
    \caption{Layer-wise probe accuracy for Llama-3.2-1B.}
    \label{fig:probe_1b}
\end{figure}

Figures~\ref{fig:probe_3b} and~\ref{fig:probe_1b} show the layer-wise linear probe accuracy for Llama-3.2-3B and Llama-3.2-1B, respectively. Similar to Llama-3-8B, the answer becomes linearly decodable at a consistent late layer across symbolic arithmetic, code, and word problems.

\subsection{Shared Arithmetic Circuits}

\begin{figure*}[t]
    \centering

    \begin{subfigure}[t]{0.49\linewidth}
        \centering
        \includegraphics[width=\linewidth]{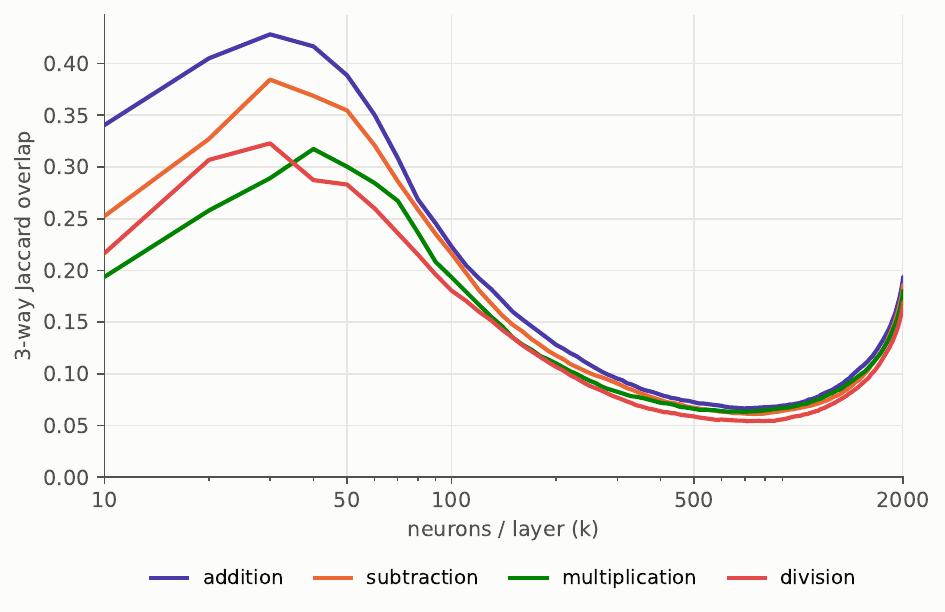}
        \caption{Llama-3.2-3B}
        \label{fig:jaccard_3b}
    \end{subfigure}
    \hfill
    \begin{subfigure}[t]{0.49\linewidth}
        \centering
        \includegraphics[width=\linewidth]{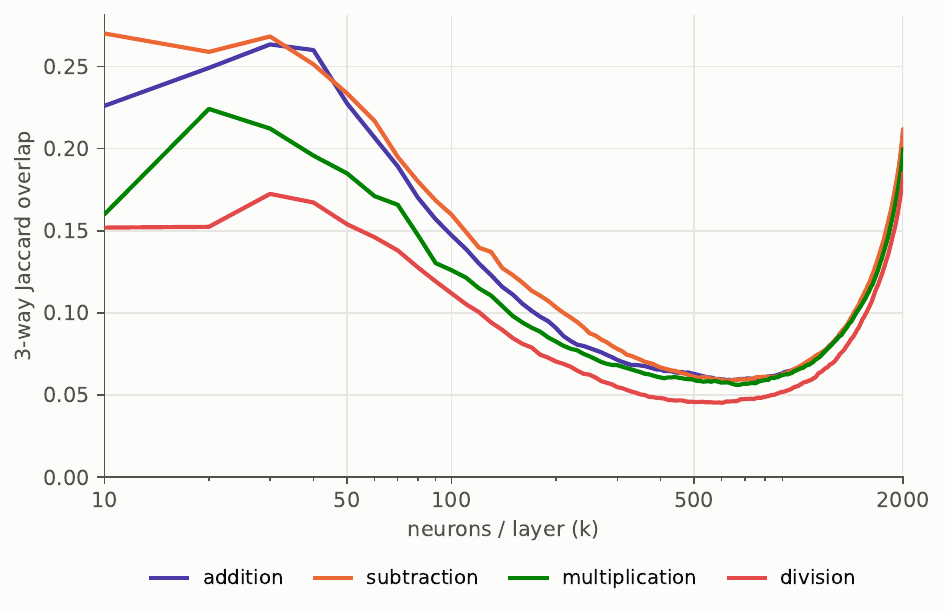}
        \caption{Llama-3.2-1B}
        \label{fig:jaccard_1b}
    \end{subfigure}

    \caption{Three-way Jaccard overlap of the top-$k$ neurons across arithmetic, code, and word formats for the additional Llama-3 models. Both models exhibit the same non-monotonic overlap pattern observed for Llama-3-8B, with peak overlap occurring at relatively small circuit sizes.}
    \label{fig:jaccard_additional}
\end{figure*}

Figures~\ref{fig:jaccard_3b} and~\ref{fig:jaccard_1b} report the three-way Jaccard overlap between arithmetic, code, and word-problem circuits. Both models exhibit the same non-monotonic overlap pattern observed for Llama-3-8B.

\subsection{Circuit Faithfulness}

\begin{figure*}[t]
    \centering
    \includegraphics[width=\linewidth]{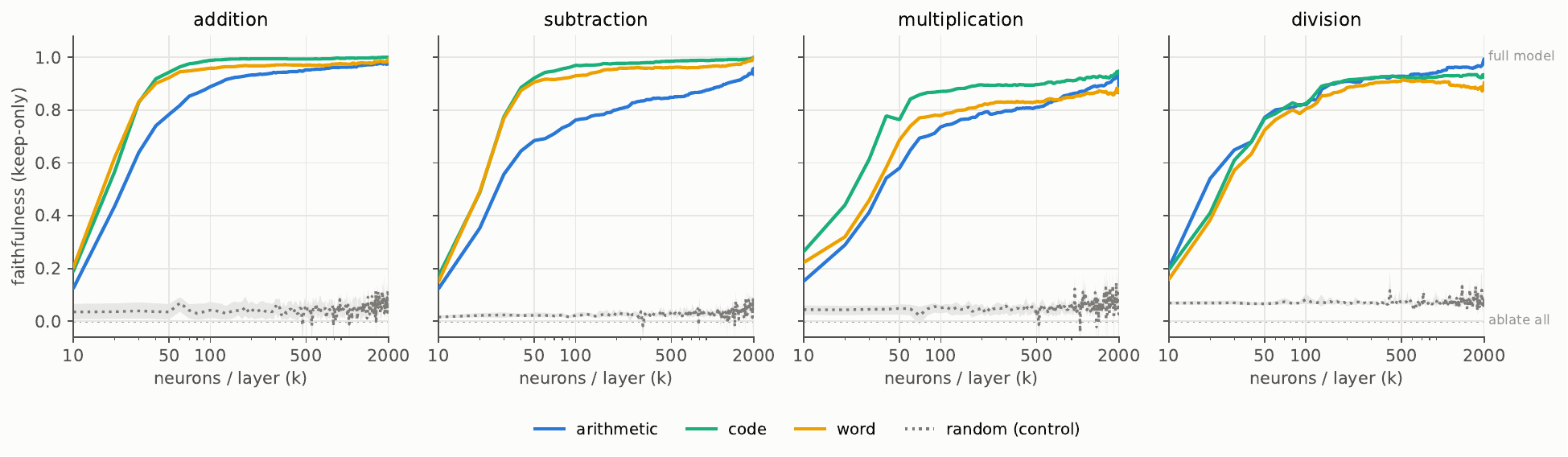}
    \caption{Keep-only intervention for Llama-3.2-3B.}
    \label{fig:keep_3b}
\end{figure*}

\begin{figure*}[t]
    \centering
    \includegraphics[width=\linewidth]{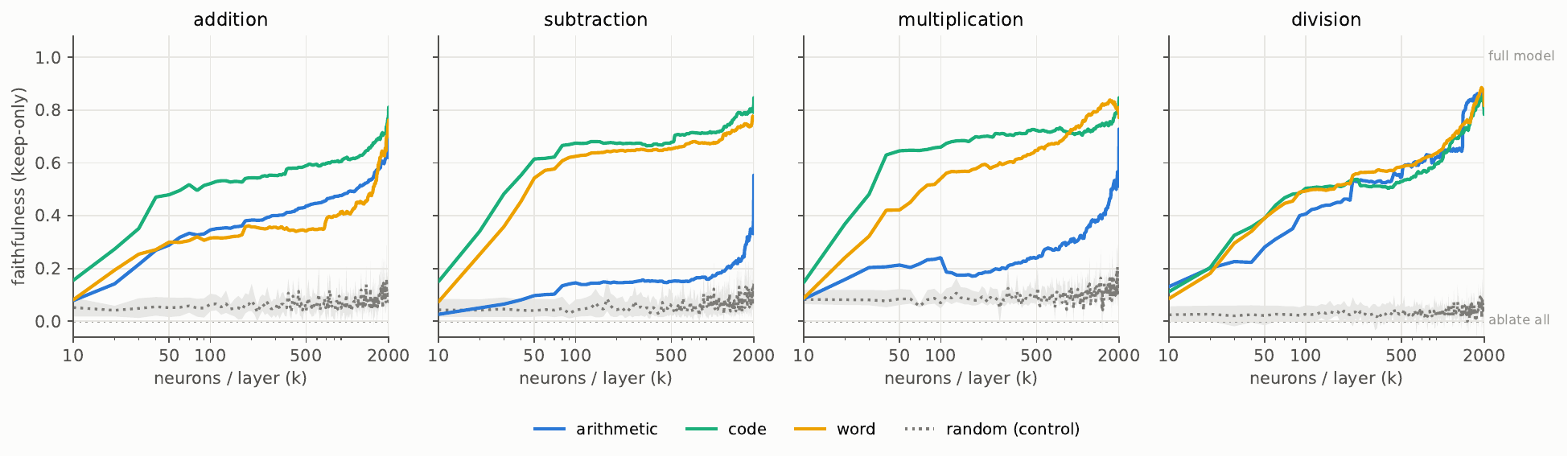}
    \caption{Keep-only intervention for Llama-3.2-1B.}
    \label{fig:keep_1b}
\end{figure*}

\begin{figure*}[t]
    \centering
    \includegraphics[width=\linewidth]{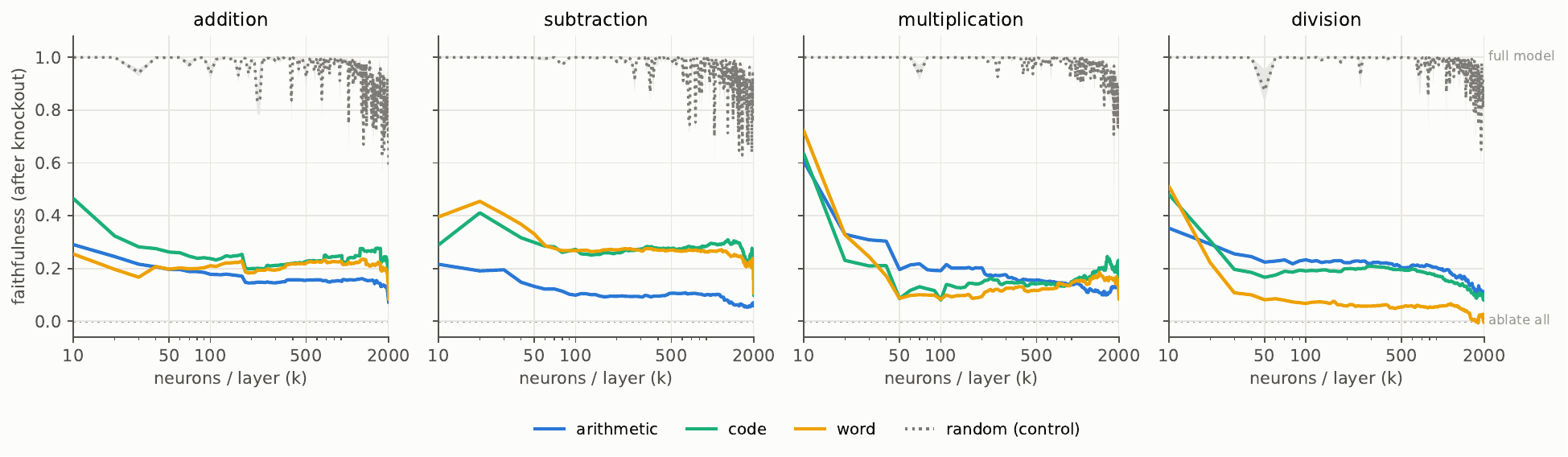}
    \caption{Knockout intervention for Llama-3.2-1B.}
    \label{fig:knock_1b}
\end{figure*}

Figures~\ref{fig:keep_3b}--\ref{fig:knock_1b} present the keep-only and knockout interventions for the two additional models. Consistent with the main paper, the shared circuit remains both necessary and sufficient for arithmetic computation.

\begin{figure*}[t]
    \centering
    \includegraphics[width=\linewidth]{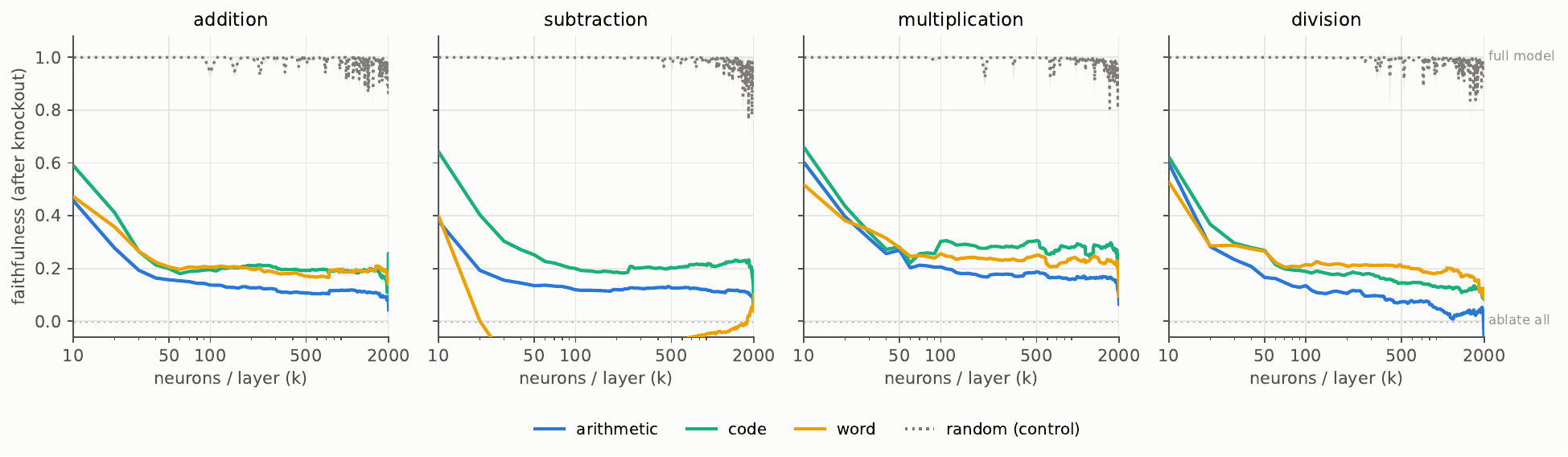}
    \caption{Knockout intervention for Llama-3.2-3B.}
    \label{fig:knock_3b}
\end{figure*}

\subsection{Cross-Format Activation Transfer}

\begin{figure*}[t]
    \centering

    \begin{subfigure}[t]{0.49\linewidth}
        \centering
        \includegraphics[width=\linewidth]{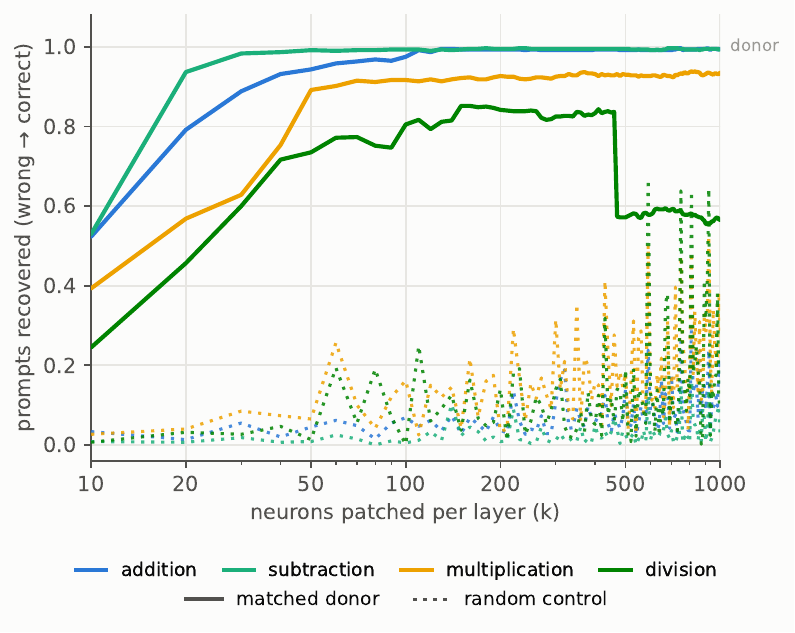}
        \caption{Code $\rightarrow$ Arithmetic}
    \end{subfigure}
    \hfill
    \begin{subfigure}[t]{0.49\linewidth}
        \centering
        \includegraphics[width=\linewidth]{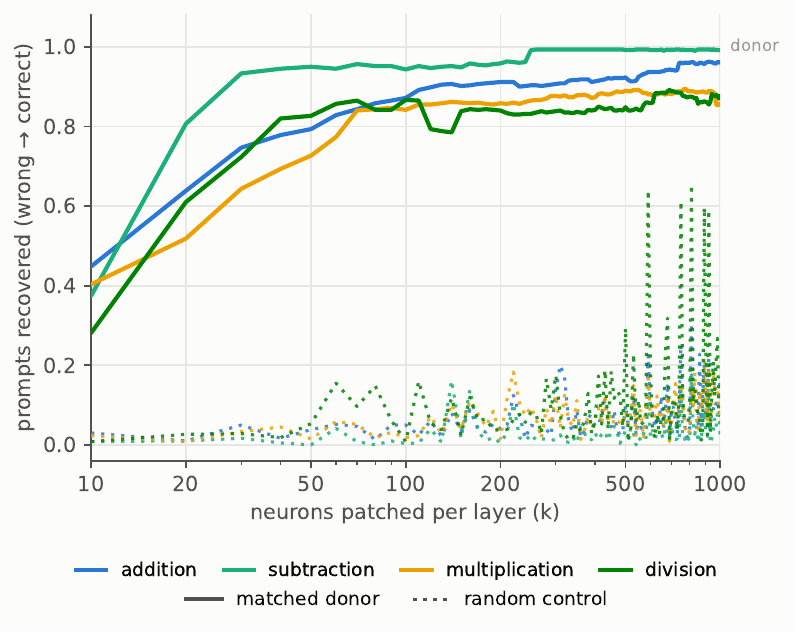}
        \caption{Word $\rightarrow$ Arithmetic}
    \end{subfigure}

    \caption{Cross-format activation transfer for Llama-3.2-3B.}
    \label{fig:xpatch_3b}
\end{figure*}

\begin{figure*}[t]
    \centering

    \begin{subfigure}[t]{0.49\linewidth}
        \centering
        \includegraphics[width=\linewidth]{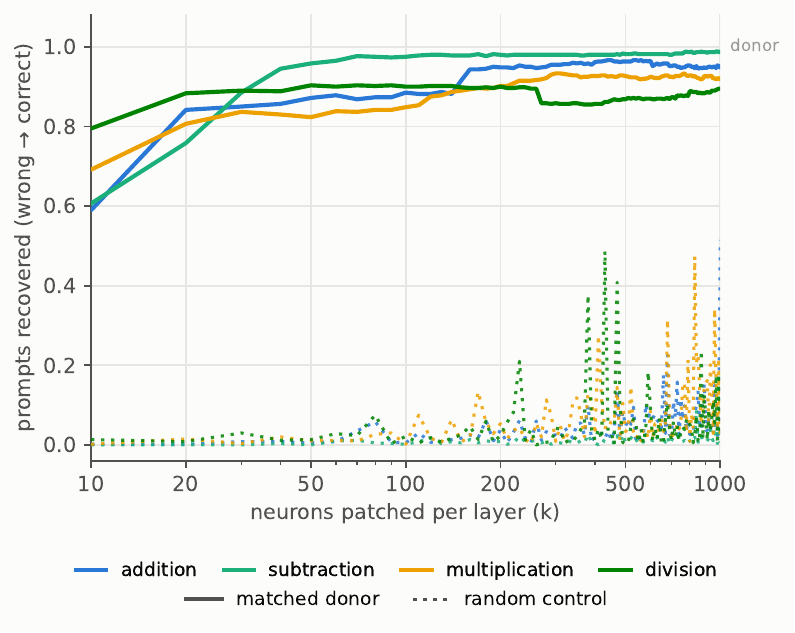}
        \caption{Code $\rightarrow$ Arithmetic}
    \end{subfigure}
    \hfill
    \begin{subfigure}[t]{0.49\linewidth}
        \centering
        \includegraphics[width=\linewidth]{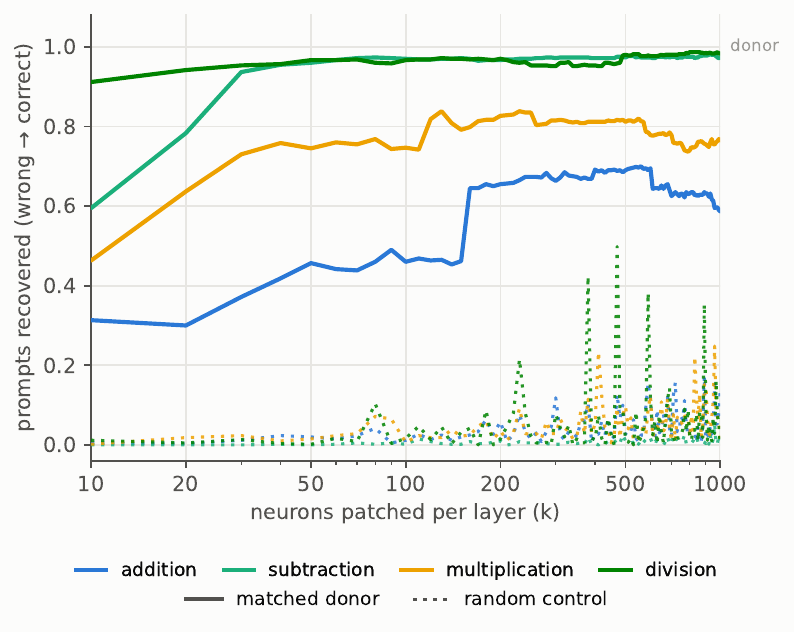}
        \caption{Word $\rightarrow$ Arithmetic}
    \end{subfigure}

    \caption{Cross-format activation transfer for Llama-3.2-1B.}
    \label{fig:xpatch_1b}
\end{figure*}

Figures~\ref{fig:xpatch_3b} and \ref{fig:xpatch_1b}  present representative cross-format activation transfer experiments for the two additional models. As in Llama-3-8B, transferring the shared heuristic-neuron activations from successful donor executions substantially improves accuracy on failed target executions.


\section{Dataset Statistics and Linear Probe Evaluation}
\label{app:dataset_statistics}

All prompts used for linear probing and circuit identification were filtered to retain only instances the models answered correctly. Table~\ref{tab:correct_census} reports the distribution of correct prompts available per model and format. While addition, subtraction, and division yielded large pools of correct executions ($\ge 20,000$), multiplication served as the primary bottleneck for dataset size due to the restricted pool of operand pairs yielding an answer $\le 999$.

Table~\ref{tab:probe_results} details the results of the format-specific, pooled-operator linear probes trained on the residual stream at the final token position. The computation onset layer (defined as the first layer where probe test accuracy exceeds 50\%) is strictly identical across all three formats within a given model.

\clearpage
\onecolumn
\begin{table}[hbp!]
\centering
\small
\begin{tabular}{llccc}
\toprule
Model & Format & Mul Correct / 2520 & Mul Accuracy & Add / Sub / Div Correct \\
\midrule
Llama-3.2-1B & Arithmetic & 528 & 21.0\% & $\ge 20,000$ \\
Llama-3.2-1B & Code & 1846 & 73.3\% & $\ge 20,000$ \\
Llama-3.2-1B & Word & 1215 & 48.2\% & $\ge 20,000$ \\
\midrule
Llama-3.2-3B & Arithmetic & 1637 & 65.0\% & $\ge 20,000$ \\
Llama-3.2-3B & Code & 2275 & 90.3\% & $\ge 20,000$ \\
Llama-3.2-3B & Word & 1236 & 49.0\% & $\ge 20,000$ \\
\midrule
Llama-3-8B & Arithmetic & 1899 & 75.4\% & $\ge 20,000$ \\
Llama-3-8B & Code & 2442 & 96.9\% & $\ge 20,000$ \\
Llama-3-8B & Word & 1645 & 65.3\% & $\ge 20,000$ \\
\bottomrule
\end{tabular}
\caption{Correct-prompt census across models and formats for answers in the range $[0, 999]$. Multiplication is the most constrained operator, as the prompt pool is exhausted at 2,520 valid pairs.}
\label{tab:correct_census}
\end{table}

\begin{table}[hbp!]
\centering
\small
\begin{tabular}{llcccccc}
\toprule
Model & Format & Total $N$ & Unique Ans & Peak Acc & Peak Layer & Onset Layer \\
\midrule
Llama-3.2-3B & Arithmetic & 9,139 & 988 & 0.891 & L26 & \textbf{L18} \\
Llama-3.2-3B & Code & 9,775 & 995 & 0.964 & L27 & \textbf{L18} \\
Llama-3.2-3B & Word & 8,736 & 994 & 0.900 & L26 & \textbf{L18} \\
\midrule
Llama-3-8B & Arithmetic & 9,399 & 990 & 0.924 & L29 & \textbf{L19} \\
Llama-3-8B & Code & 9,943 & 995 & 0.967 & L29 & \textbf{L19} \\
Llama-3-8B & Word & 9,144 & 994 & 0.933 & L29 & \textbf{L19} \\
\midrule
Llama-3.2-1B & Arithmetic & 8,028 & 931 & 0.733 & L15 & \textbf{L14} \\
Llama-3.2-1B & Code & 9,346 & 992 & 0.784 & L15 & \textbf{L14} \\
Llama-3.2-1B & Word & 8,712 & 979 & 0.695 & L15 & \textbf{L14} \\
\bottomrule
\end{tabular}
\caption{Linear probe evaluation across formats. The onset layer represents the first transformer layer where test accuracy exceeds 0.50, demonstrating identical computation onset points across formats.}
\label{tab:probe_results}
\end{table}

\twocolumn
\section{Additional Heuristic Consistency Results}
\label{app:heuristic_consistency}
Table~\ref{tab:app_detailed_heuristics} reports the complete heuristic consistency statistics for Llama-3-8B, Llama-3.2-3B and Llama-3.2-1B. The trends observed in the main paper are consistent across models: pairwise and three-way heuristic agreement remain high for all arithmetic operators, although exact family agreement is more stringent. Division exhibits lower agreement, likely reflecting the smaller and less diverse set of valid integer quotients.

\section{Distribution of Heuristic Families}
\label{app:family_distribution}


To provide a granular understanding of the shared circuit's composition, we expand the broad heuristic categories introduced in Section 3.2 into specific functional families. Following the algorithmic implementation of \citet{nikankinArithmeticAlgorithmsLanguage2025}, the classification process evaluates neurons based on both a specific \textit{scoring mechanism} (range, modulo, pattern, or exact value) and a specific \textit{target} (a single operand, both operands, or the expected result). 

In our analysis, we group the parameterized heuristic labels into distinct functional families defined by the intersection of these targets and scorers. This yields families such as \textbf{operand range}, \textbf{result range}, \textbf{both modulo}, and \textbf{result value} (which activates for exact, zero-padded target numbers). We additionally include the \textbf{identical operands} ($op_1 = op_2$) heuristics.
Table \ref{tab:dist_llama3_8b}, \ref{tab:dist_llama32_3b}, and \ref{tab:dist_llama32_1b} summarizes the distribution of these heuristic families for the shared circuit. 

\begin{table}[ht]
\centering
\begin{tabular}{lrrr}
\toprule
\textbf{Heuristic Family} & \textbf{Arithmetic} & \textbf{Code} & \textbf{Word} \\
\midrule
Result Pattern & 808 & 800 & 758 \\
Result Modulo & 596 & 606 & 558 \\
Result Range & 387 & 389 & 411 \\
Operand Pattern & 222 & 226 & 292 \\
Operand Range & 167 & 96 & 175 \\
Both Range & 63 & 61 & 61 \\
Both Modulo & 22 & 34 & 40 \\
Result Value & 22 & 28 & 32 \\
Operand Modulo & 17 & 11 & 19 \\
Operand Value & 0 & 1 & 3 \\
Identical Operands & 0 & 6 & 4 \\
\bottomrule
\end{tabular}
\caption{Family distribution of shared neurons for Llama-3-8B (addition). A single neuron may be assigned multiple families.}
\label{tab:dist_llama3_8b}
\end{table}

\begin{table}[ht]
\centering
\begin{tabular}{lrrr}
\toprule
\textbf{Heuristic Family} & \textbf{Arithmetic} & \textbf{Code} & \textbf{Word} \\
\midrule
Result Pattern & 631 & 691 & 643 \\
Result Modulo & 490 & 552 & 536 \\
Result Range & 247 & 350 & 312 \\
Operand Range & 189 & 74 & 119 \\
Operand Pattern & 170 & 147 & 202 \\
Both Range & 54 & 55 & 56 \\
Both Modulo & 15 & 20 & 24 \\
Operand Modulo & 11 & 8 & 16 \\
Result Value & 5 & 32 & 23 \\
Identical Operands & 0 & 2 & 5 \\
\bottomrule
\end{tabular}
\caption{Family distribution of shared neurons for Llama-3.2-3B (addition). A single neuron may be assigned multiple families.}
\label{tab:dist_llama32_3b}
\end{table}

\begin{table}[ht]
\centering
\begin{tabular}{lrrr}
\toprule
\textbf{Heuristic Family} & \textbf{Arithmetic} & \textbf{Code} & \textbf{Word} \\
\midrule
Result Modulo & 166 & 183 & 175 \\
Operand Pattern & 147 & 160 & 228 \\
Result Pattern & 124 & 160 & 87 \\
Result Range & 92 & 151 & 71 \\
Operand Range & 78 & 45 & 129 \\
Operand Modulo & 32 & 22 & 36 \\
Both Range & 20 & 32 & 12 \\
Both Modulo & 18 & 16 & 19 \\
Operand Value & 0 & 3 & 7 \\
Result Value & 0 & 1 & 0 \\
Identical Operands & 0 & 0 & 1 \\
\bottomrule
\end{tabular}
\caption{Family distribution of shared neurons for Llama-3.2-1B (addition). A single neuron may be assigned multiple families.}
\label{tab:dist_llama32_1b}
\end{table}


\section{Statistical Significance of Heuristic Agreement}
\label{app:heuristic_significance}

To verify that the functional consistency of shared neurons exceeds random chance, we established a null distribution by randomly permuting the heuristic family assignments within each format independently. Table~\ref{tab:heuristic_significance} reports the observed exact family match rates and mean pairwise Jaccard similarities against the null expectations. Across all models and operators, the observed functional agreement significantly exceeds the maximum values observed under the null distribution ($p < 10^{-4}$).

\clearpage

\clearpage
\onecolumn
\begin{table}[hbp!]
\centering
\small
\begin{tabular}{llrrrrrr}
\toprule
model & operator & $|C|$ & labelled & 3-way common & exact match & Jaccard (pair) & Jaccard (3-way) \\
\midrule
Llama-3-8B & addition & 784 & 594 & 96.0\% & 53.9\% & 0.83 & 0.75 \\
Llama-3-8B & subtraction & 803 & 761 & 96.8\% & 23.5\% & 0.73 & 0.62 \\
Llama-3-8B & multiplication & 733 & 733 & 99.9\% & 20.1\% & 0.76 & 0.65 \\
Llama-3-8B & division & 768 & 424 & 88.0\% & 24.8\% & 0.67 & 0.54 \\
\midrule
Llama-3.2-3B & addition & 661 & 492 & 94.5\% & 44.3\% & 0.79 & 0.70 \\
Llama-3.2-3B & subtraction & 618 & 566 & 96.8\% & 14.5\% & 0.68 & 0.55 \\
Llama-3.2-3B & multiplication & 573 & 573 & 99.8\% & 15.2\% & 0.74 & 0.63 \\
Llama-3.2-3B & division & 574 & 286 & 75.5\% & 18.5\% & 0.59 & 0.44 \\
\midrule
Llama-3.2-1B & addition & 332 & 203 & 85.2\% & 36.0\% & 0.70 & 0.58 \\
Llama-3.2-1B & subtraction & 372 & 354 & 96.6\% & 7.9\% & 0.64 & 0.50 \\
Llama-3.2-1B & multiplication & 299 & 299 & 100.0\% & 16.7\% & 0.75 & 0.64 \\
Llama-3.2-1B & division & 267 & 168 & 69.6\% & 6.0\% & 0.50 & 0.31 \\
\bottomrule

\end{tabular}
\caption{Per-operator heuristic consistency statistics for all models. $|C|$ denotes the size of the shared circuit, \emph{labelled} denotes the number of neurons assigned a heuristic family, \emph{3-way common} denotes the percentage of neurons receiving a heuristic assignment in all three formats, \emph{Exact} denotes the percentage assigned the same heuristic family across all formats, and Pair/3-way denote the corresponding heuristic-family Jaccard similarities.}
\label{tab:app_detailed_heuristics}
\end{table}

\begin{table*}[hbp!]
\centering
\small
\begin{tabular}{llrrrrrrrr}
\toprule
& & \multicolumn{4}{c}{exact family match} & \multicolumn{4}{c}{mean pairwise Jaccard} \\
\cmidrule(lr){3-6}\cmidrule(lr){7-10}
model & operator & obs & null & null max & $p$ & obs & null & null max & $p$ \\
\midrule
Llama-3-8B & addition & 0.539 & 0.052 & 0.087 & $\!<1e-4$ & 0.700 & 0.315 & 0.339 & $\!<1e-4$ \\
Llama-3-8B & subtraction & 0.235 & 0.009 & 0.021 & $\!<1e-4$ & 0.707 & 0.405 & 0.421 & $\!<1e-4$ \\
Llama-3-8B & multiplication & 0.201 & 0.024 & 0.048 & $\!<1e-4$ & 0.757 & 0.568 & 0.582 & $\!<1e-4$ \\
Llama-3-8B & division & 0.248 & 0.021 & 0.061 & $\!<1e-4$ & 0.488 & 0.183 & 0.207 & $\!<1e-4$ \\
\midrule
Llama-3.2-3B & addition & 0.443 & 0.048 & 0.089 & $\!<1e-4$ & 0.658 & 0.298 & 0.319 & $\!<1e-4$ \\
Llama-3.2-3B & subtraction & 0.145 & 0.008 & 0.027 & $\!<1e-4$ & 0.641 & 0.383 & 0.402 & $\!<1e-4$ \\
Llama-3.2-3B & multiplication & 0.152 & 0.018 & 0.042 & $\!<1e-4$ & 0.743 & 0.567 & 0.581 & $\!<1e-4$ \\
Llama-3.2-3B & division & 0.185 & 0.017 & 0.054 & $\!<1e-4$ & 0.414 & 0.151 & 0.179 & $\!<1e-4$ \\
\midrule
Llama-3.2-1B & addition       & 0.360                                  & 0.042                                     & 0.097    & $\!<1e-4$ & 0.510 & 0.226 & 0.257    & $\!<1e-4$ \\
Llama-3.2-1B & subtraction    & 0.079                                  & 0.014                                     & 0.040    & $\!<1e-4$ & 0.623 & 0.409 & 0.431    & $\!<1e-4$ \\
Llama-3.2-1B & multiplication & 0.167                                  & 0.021                                     & 0.057    & $\!<1e-4$ & 0.749 & 0.579 & 0.597    & $\!<1e-4$ \\
Llama-3.2-1B & division       & 0.060                                  & 0.007                                     & 0.041    & $\!<1e-4$ & 0.392 & 0.189 & 0.219    & $\!<1e-4$ \\
\bottomrule
\end{tabular}
\caption{Statistical significance of heuristic family agreement. Observed (\emph{obs}) exact match rates and pairwise Jaccard similarities are compared against a null distribution generated via random permutation of heuristic labels. Across all settings, the observed agreement significantly exceeds chance ($p < 10^{-4}$).}
\label{tab:heuristic_significance}
\end{table*}

\end{document}